\documentclass[runningheads]{llncs}
\usepackage{graphicx}
\begin{document}
\title{BWCNN: Blink to Word, a Real-Time Convolutional Neural Network Approach
}
\titlerunning{ }
%
\author{Albara Ah Ramli\inst{1}
\and
Rex Liu\inst{1}
\and
Rahul Krishnamoorthy\inst{2}
\and
Vishal I B\inst{2}
\and
Xiaoxiao Wang\inst{1}
\and
Ilias Tagkopoulos\inst{1}
\and
Xin Liu\inst{1}
}

\authorrunning{ }
%
\institute{
Department of Computer Science, University of California, Davis, CA, USA\\
\and 
Electrical and Computer Engineering, University of California, Davis, CA, USA\\
\email{\{arramli, rexliu, rkrishnamoorthy, vib, xxwa, iliast, xinliu\}@ucdavis.edu}}
\maketitle 
%
\begin{abstract}
Amyotrophic lateral sclerosis (ALS) is a progressive neurodegenerative disease of the brain and the spinal cord, which leads to paralysis of motor functions. Patients retain their ability to blink, which can be used for communication. Here, We present an Artificial Intelligence (AI) system that uses eye-blinks to communicate with the outside world, running on real-time Internet-of-Things (IoT) devices. The system uses a Convolutional Neural Network (CNN) to find the blinking pattern, which is defined as a series of Open and Closed states. Each pattern is mapped to a collection of words that manifest the patient's intent. To investigate the best trade-off between accuracy and latency, we investigated several Convolutional Network architectures, such as ResNet, SqueezeNet, DenseNet, and InceptionV3, and evaluated their performance. We found that the InceptionV3 architecture, after hyper-parameter fine-tuning on the specific task led to the best performance with an accuracy of 99.20\% and 94ms latency. This work demonstrates how the latest advances in deep learning architectures can be adapted for clinical systems that ameliorate the patient's quality of life regardless of the point-of-care.

\end{abstract}

\section{Introduction}

A plethora of clinical conditions, such as brain trauma and amyotrophic lateral sclerosis (ALS), cause damage to the central neural system (CNS) or brain, in such as way that the ability of speech and motor functions cannot be sustained. In those cases, the ability to communicate is limited, to non-verbal forms of communication, such as eye blinking. 

Certain approaches have been used in the past to solve this problem. Researchers in \cite{b1} use Infrared (IR) sensors to estimate the state of the eyes ({\lq}Open{\rq} or {\lq}Closed{\rq}) in order to detect blinking, which is then converted to Morse code. Challenges in this approach include the IR sensor being irradiated by other sources resulting in false eye-blinks, and the risk of cataract formation in the case of prolonged use. \cite{b1}. In another study, a traditional computer vision-based system that detects users' eye-blinks, measures their duration, and interprets the blinks in real-time was proposed \cite{b2}. Although the system has achieved an accuracy of 95.6\% in ideal conditions, traditional image processing techniques are prone to failure under limited lighting conditions, arbitrary changes in image texture, changes in users pose, among others, which limits their real-time accuracy, and make these systems not robust in real applications.

A wearable device to detect eye-blinks for alleviating dry eyes was proposed in \cite{b8}. This method captured 85.2\% of all the blinks that occurred during testing. But, the IR sensors tend to show false readings when the orientations are altered and hence, are unreliable in realistic scenarios. Any facial movements such as laughing, or yawning can induce errors. 

To avoid these shortcomings, we here present a deep-learning vision system that detects eye-blinks and maps them to words in real-time. Our system uses the InceptionV3 \cite{b13} architecture and achieves an accuracy of 99.20\% with a latency of 94.1ms on IoT devices. Our method is safe to use, has better performance than previous methods, it is robust to changes in lighting conditions and facial orientation, and its architecture is modular so its output can be mapped to other tasks, such as controlling software applications or devices. The main contribution of this paper resides in designing, implementing, and evaluating the first deep-learning solution for eye-blink communication with performance, latency, and safety specifications that can be used in a real-world environment. 

\begin{figure}[htbp] 
\centerline{\includegraphics[width=0.7\textwidth]{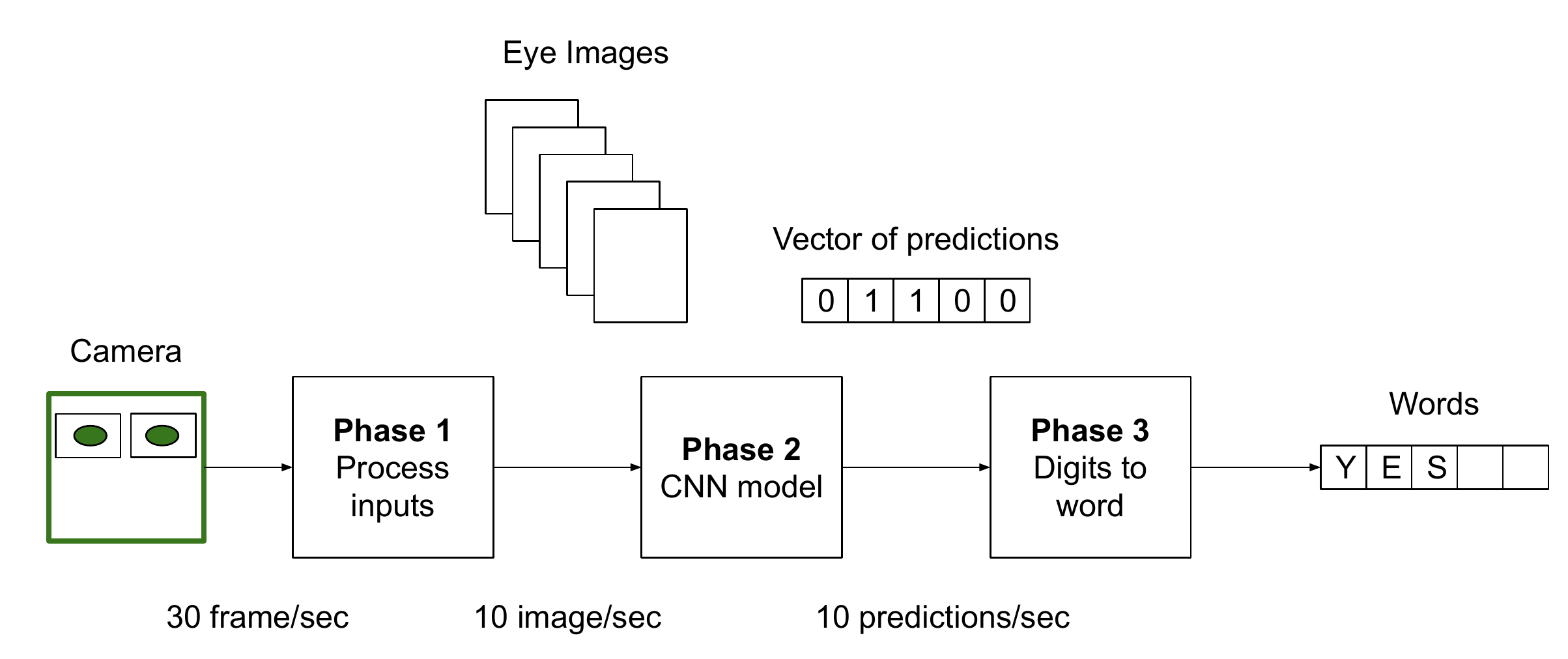}}
\caption{The 3 phases of the BWCNN system.}
\label{fig_hld}
\end{figure}

\subsection{Previous work}
An efficient system for eye-blink detection is presented in \cite{b3}. This method uses Haar-cascade classifiers for face detection and eye positions. The performance of Haar-cascade classifiers is not invariant to the change in lighting conditions. Hence there is a decay in performance. 

A low-cost implementation of an eye-blink-based communication aid for ALS patients is presented in \cite{b4}. Template matching is used to track the eye and detect eye-blinks using hierarchical optical flow. The implementation has an accuracy of 94.75\% during the typing test. However, the algorithm takes approximately 2 seconds to generate a single scan of the eye. This is excessive for a single character. 

This paper \cite{b7} presents a real-time detection and classification between eye-blink (with both eyes), left wink, or right wink. with 96, 92, and 88\% accuracy. The latency for the detection of a single blink was 250ms.

A wearable device to detect eye-blinks for alleviating dry eyes was proposed in \cite{b8}. This method captured 85.2\% of all the blinks that occurred during testing. But, the IR sensors tend to show false readings when the orientations are altered and hence, are unreliable in realistic scenarios. Any facial movements such as laughing, or yawning can induce errors. 

\section{Design and Implementation}
Our system detects the state of eyes, {\lq}Open{\rq} and {\lq}Closed{\rq}, even under poor lighting conditions. We have a pre-defined set of patient inputs corresponding to the blink pattern, which we map to actions in real-time. These inputs could correspond to movements (up, down, left, right), clicks, etc., which would enable patients to use different applications (browser, email, etc.) or devices (mouse, keyboard, etc). As a proof-of-concept, we mapped these inputs to specific words. Since we use predefined words instead of Morse code or other encoding patterns, it is simple for patients to spell out a sentence. We want to create a system that works almost flawlessly in real-time and is safe to use. This is represented by equation (1). \(\mathcal{P}\) refers to the performance of the system, which in our case is the accuracy. 
\(\mathcal{S}\) refers to the System parameters. 
\(\mathcal{W}\) refers to the weights of the neural network. 
\(\mathcal{A}\) refers to the architecture.
\(\mathcal{H}\) is the set of all architectures that can be used for this purpose. 
\(\mathcal{T}\) refers to time constraint which is the prediction time (response time, or latency) for the model on the validation set. It is important to reduce the response time of the system since it will be running over a real-time IoT device such as Raspberry Pi. 
\noindent\\
\begin{equation}
\max_{\mathcal{A} \in \mathcal{H}} \mathcal{P} ( \mathcal{S} , \mathcal{W} , \mathcal{A} ) 
\end{equation}
$\;\;\;\;\;\;\;\;\;\;\;\;\;\;\;\;\;\;\;\;\;\;\;\;\;\;\;\;\;\;\;\;\;\;\;\;\;\;\;\;\;\;\;\;\;\;\;\;\;\;s.t\;prediction\;time\leq \mathcal{T}$\\

The goal of this approach is to detect if the patients blink their eyes and to map the sequence of blinks to a particular entry in a dictionary of words. In order to achieve this, one has to detect the state of eyes ({\lq}Open{\rq} or {\lq}Closed{\rq}). If an {\lq}Open{\rq} state is followed by a {\lq}Closed state, the system detects an eye-blink, as shown in Fig.~\ref{fig_B-good}. The system is divided into three phases as shown in Fig.~\ref{fig_hld}. 

\subsection{Phase 1: Capturing and saving a stream of frames}
In this section, we present the method of obtaining the data and preprocessing it to be given as input to the ConvNet. The system uses a camera device attached to IoT for capturing the frames. Regular webcams are capable of capturing about 30 frames per second (fps). The fps will directly affect the user experience. A higher fps will increase the latency. 

Since the system runs in real-time, it is more effective to reduce the latency. At the same time, using lower fps can lead to missing a blink. Our experiment shows that 10 frames per second are a reasonable frame rate for real-time application. As there is a frame being captured every 100ms, our model must predict each frame in less than 100ms to avoid delay.
We further impose a constraint on the user that the {\lq}Closed{\rq} state should be maintained for at least 200ms. The system saves each frame as a gray-scale image of dimensions $80\times70$ pixels. Each image is 2KB in size. 

\subsection{Phase 2: Predict the content of the image}
In this section, we present the experiments to choose the best fitting neural network architecture to predict the state of the eye. We compare four state-of-the-art architectures (SqueezeNet ~\cite{b5}, ResNet ~\cite{b9}, InceptionV3 and DenseNet architecture). Both the architecture and the hyperparameters play a large role in model performance. We start by training the networks from scratch for different batch sizes and find the batch size that gives the best results. We further explore the chances of improving performance by using transfer learning.

\subsubsection{Training Dataset:} We used the eye dataset from Media Research Lab (MRL). The dataset contains 84,898 images of eyes taken from 37 individuals consisting of 33 men and 4 women. Each image in the dataset was collected from one of the following sensors: Intel RealSense RS 300 sensor with a resolution of $640\times480$, IDS Imaging sensor with a resolution of $1280\times1024$, and Aptina sensor with a resolution of $752\times480$. The original dataset contains 6  classes: {\lq}gender{\rq}, {\lq}glasses{\rq}, {\lq}eye state{\rq}, {\lq}reflections{\rq}, {\lq}lighting conditions{\rq}, and {\lq}sensor resolution{\rq}. We split the dataset into an 80\% training set and a 20\% test set. 

\subsubsection{Training experiments:}
Two important considerations when training the model were accuracy and latency. Latency for detection is the time taken to make an accurate classification. To find the model which can provide the best accuracy with the least latency, we implemented SqueezeNet, ResNet, InceptionV3, and DenseNet.

\begin{table*}[t]
\caption{ResNet with and without transfer learning}
\label{RESNET}
\begin{center}
\begin{tabular}{|c|c|c|c|c|c|c|c|c|}
\hline
\textbf{} & \textbf{} & \multicolumn{2}{c|}{\textbf{
\begin{tabular}[c]{@{}c@{}}Training a model\\ (from scratch)\end{tabular}
}} & \multicolumn{2}{c|}{\textbf{
\begin{tabular}[c]{@{}c@{}}Transfer learning \\ (from our  model)\end{tabular}
}} & \multicolumn{2}{c|}{
\textbf{
\begin{tabular}[c]{@{}c@{}}Transfer learning \\ (from official)\end{tabular}
}
} \\ \hline
\textbf{Batch S.} & 
\textbf{Epoch} & 
\textbf{Acc. (\%)} & 
\textbf{Ep.imp.} & 
\textbf{Acc. (\%)} & 
\textbf{Ep. imp.} & 
\textbf{Acc. (\%)} & 
\textbf{Ep. imp.} \\ \hline
8 & 100 & 99.21 & 32 & 99.22 & 31 & 99.22 & 29 \\ \hline
16 & 100 & 99.26 & 55 & 99.23 & 51 & 99.17 & 16\\ \hline
32 & 100 & 99.22 & 48  & 99.22 & 49 & 99.17 & 25\\ \hline
\end{tabular}
\end{center}
\end{table*}

\begin{table*}[t]
\caption{DenseNet,SqueezeNet,InceptionV3}
\label{DenseNet,SqueezeNet,InceptionV3}
\begin{center}
\begin{tabular}{|c|c|c|c|c|c|c|c|}
\hline
\textbf{} & \textbf{} & \multicolumn{2}{c|}{\textbf{DENSENET}} & \multicolumn{2}{c|}{\textbf{SQUEEZENET}} & \multicolumn{2}{c|}{\textbf{INCEPTIONV3}} \\ \hline
\textbf{Batch S.} & \textbf{Eep.} & \textbf{Acc. (\%)} & \textbf{Ep. imp.} & \textbf{Acc. (\%)} & \textbf{Ep. imp.} & \textbf{Acc. (\%)} & \textbf{Ep. imp.} \\ \hline
8 & 100 & 99.24 & 55 & 49.40 & 1 & 99.14 & 35 \\ \hline
16 & 100 & 99.18 & 70 & 49.40 & 1 & 99.20 & 22 \\ \hline
32 & 100 & 99.21 & 52 & 49.40 & 1 & 99.17 & 38 \\ \hline
\end{tabular}
\end{center}
\end{table*}

\subsubsection{Train ResNet architecture from scratch:}
We trained the ResNet architecture for 100 epochs using 6 batch sizes. The performance metric used is the overall accuracy. Comparing the performance, we selected the three best batch sizes from 1, 2, 4, 8, 16, and 32.  Table.~\ref{RESNET} shows that batch sizes 8, 16, and 32 provide the best accuracy. Since our experiment stops at 100 epochs, training the network for more epochs might improve the performance. To test this, we ran ResNet, for all batch sizes for 500 epochs. The results show that there is no further improvement in accuracy.
\subsubsection{Transfer learning:}
We took the weights from the most accurate architecture, ResNet with batch size 16, and transferred these weights to all other ResNet architecture with the 3 best performing batch sizes. We also did transfer learning by using weights from the official pre-trained ResNet on our ResNet models with the 3 best performing batch sizes. As before, after 100 epochs, there was no significant improvement in the accuracy of the network. 

\subsubsection{Train InceptionV3, SqueezeNet and DeneseNet architectures from scratch:}
We assume that the 3 best-performing batch sizes from ResNet would be the best performing in the other architectures as well. To investigate this assumption we run the same experiment again but with DenseNet, Inception, and SqueezeNet as shown in Table.~\ref{DenseNet,SqueezeNet,InceptionV3}.

\subsection{Phase 3: Mapping}
\begin{figure}[htbp] 
\centerline{\includegraphics[width=0.9\textwidth]{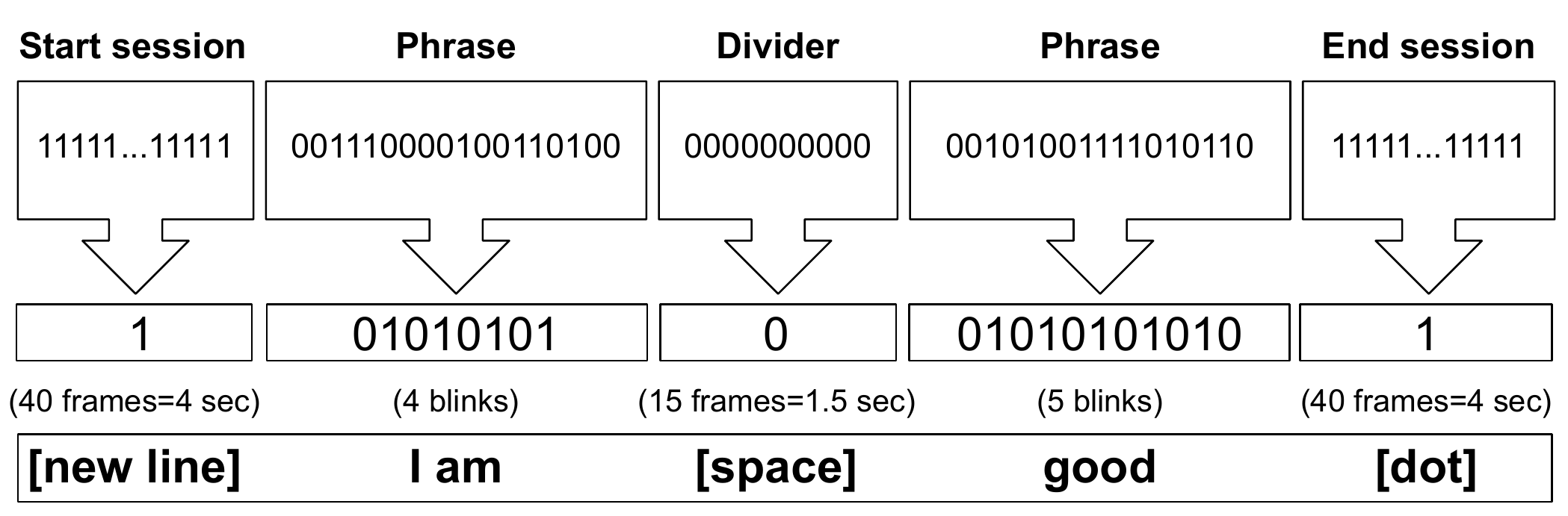}}
\caption{BWCNN system predicting a series of frames in the real-time}
\label{fig_B-good}
\end{figure}

Our system stored the output of the neural network as a vector of 0s and 1s, where Zero represents the {\lq}Open{\rq} state for the eye and a One represents the {\lq}Closed{\rq} state as shown in Fig.~\ref{fig_B-good}. We normalize the vector by truncating repeated instances of a state with a single instance. For example, vector 00000110000 becomes 010. The system provides a special word to start and end a session of blinking, which is keeping the eyes closed for 4 seconds. 
The system recognizes the end of a word when the patient's eyes remain open for 1 second.
Based on the output vector, the number of blinks is calculated and is mapped with the words in the dictionary (Table.~\ref{Dictionary}). The dictionary consists of basic words that we use as a proof-of-concept and it can be modified.

\begin{table}
\caption{Dictionary}
\label{Dictionary}
\begin{center}
\begin{tabular}{|c|c|c|c|c|c|c|c|}
\hline
\textbf{\# blinks} & \textbf{1} & \textbf{2} & \textbf{3} & \textbf{4} & \textbf{5} & \textbf{6} & \textbf{7} \\ \hline
\textbf{Pattern} & 1 & 101 & 10101 & 1010101 & 101010101 & 10101010101 & 1010101010101 \\ \hline
\textbf{Words} & Yes & No & Hi & I am & Good & Thanks & How are you? \\ \hline
\textbf{Mouse} & Right & Left & Click R. & Click L. & Up & Down & Hold click \\ \hline
\textbf{Keyboard} & Tab & Enter & Back & Esc & Scroll up & Scroll down & Space \\ \hline
\end{tabular}
\end{center}
\end{table}

\section{Results}
This section provides the performance results of the different architectures. Table.~\ref{RESNET} lists the results obtained from training ResNet for 100 epochs on batch sizes 1, 2, 4, 8, 16, and 32. Training ResNet from scratch for 100 epochs on a batch size of 16 yields the best accuracy, 99.26\%. There is no improvement in accuracy after 55 epochs. 

\begin{table}[htbp]
\caption{Final results: 3.1 GHz Quad-Core Intel Core i7, 16 GB 2,133 MHz} 
\label{Final Results}
\begin{tabular}{|c|c|c|c|c|c|c|c|c|}
\hline
\textbf{\begin{tabular}[c]{@{}c@{}}Model\\ Archit-\\ ecture\end{tabular}} & \textbf{\begin{tabular}[c]{@{}c@{}}Batch \\ size\end{tabular}} & \textbf{\begin{tabular}[c]{@{}c@{}}Ep. \\ imp.\end{tabular}} & \textbf{\begin{tabular}[c]{@{}c@{}}Total \\ params\end{tabular}} & \textbf{\begin{tabular}[c]{@{}c@{}}Trainable \\ params\end{tabular}} & \textbf{\begin{tabular}[c]{@{}c@{}}Non-\\ trainable \\ params\end{tabular}} & \textbf{\begin{tabular}[c]{@{}c@{}}Model \\ size\end{tabular}} & \textbf{\begin{tabular}[c]{@{}c@{}}Model\\ Accu-\\ racy\end{tabular}} & \textbf{\begin{tabular}[c]{@{}c@{}}Avg \\ latency\end{tabular}} \\ \hline
ResNet & 16 & 55 & 23,591,810 & 23,538,690 & 53,120 & 283MB & 99.26\% & 117.28ms \\ \hline
DenseNet & 8 & 55 & 7,039,554 & 6,955,906 & 83,648 & 85MB & 99.24\% & 146.09ms \\ \hline
SqueezeNet & 16 & 1 & 723,522 & 723,522 & 0 & 8MB & 49.40\% & 13.64ms \\ \hline
InceptionV3 & 16 & 16 & 21,806,882 & 1,772,450 & 34,432 & 262MB & 99.20\% & 94.1ms \\ \hline
\end{tabular}
\end{table}

\begin{figure} 
\centerline{\includegraphics[width=0.9\textwidth]{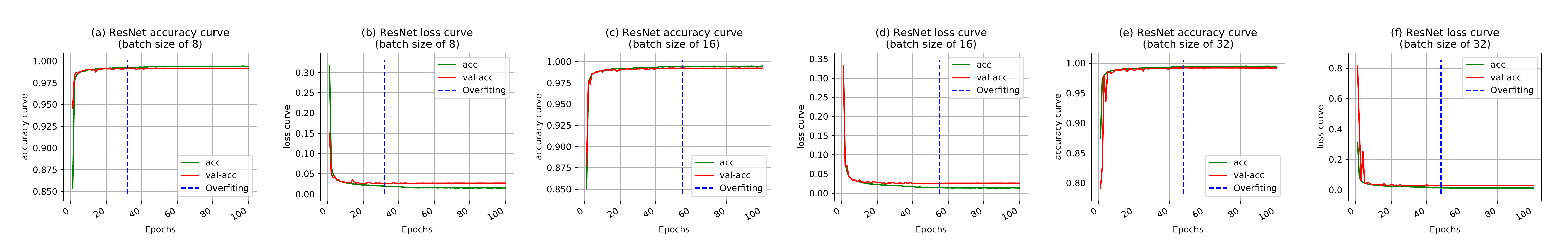}}
\caption{ResNet: training and validation (accuracy and loss) curves.}
\label{fig_pdf_ResNet}
\end{figure}

\begin{figure} 
\centerline{\includegraphics[width=0.9\textwidth]{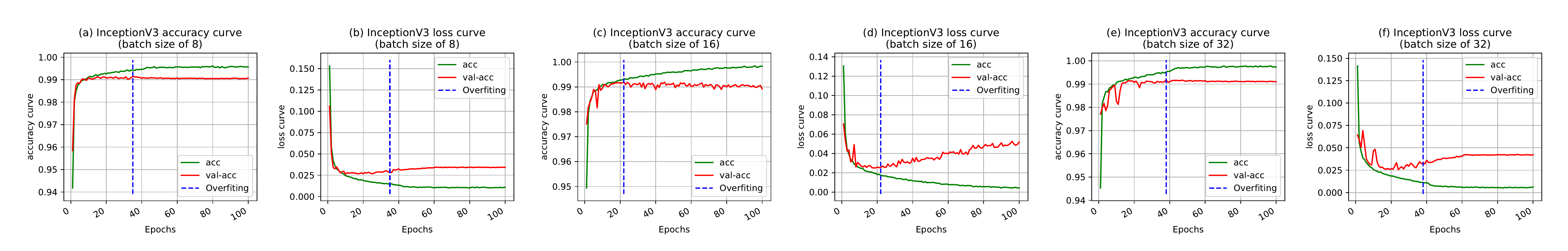}}
\caption{InceptionV3: training and validation (accuracy and loss) curves.}
\label{fig_pdf_InceptionV3}
\end{figure}

We trained a new ResNet model using transfer learning. We took the weights from the best performing network that we had trained form scratch and used them to train a ResNet network with batch sizes of 8, 16, and 32. We also did transfer learning with weights obtained from the official pre-trained ResNet architecture and used it to train ResNet architecture with batch sizes of 8, 16, and 32. The results are shown in Table.~\ref{RESNET}.

Table.~\ref{DenseNet,SqueezeNet,InceptionV3} presents the results for DenseNet, SqueezeNet, and InceptionV3 architectures for the batch sizes of 8, 16, and 32.
Fig.~\ref{fig_pdf_InceptionV3} shows the training and loss curve for InceptionV3. DenseNet has the best accuracy of 99.24\%, which is achieved for a batch size of 8, as training for 55 epochs, after which it starts over-fitting.

SqueezeNet seems to be the worst among all the architectures with an accuracy of 49.40\%. There is no change in accuracy after the first epoch, which is in Table.~\ref{DenseNet,SqueezeNet,InceptionV3}. The final comparison of the best results across the different architectures in Table.~\ref{Final Results}. Apart from the accuracy of the networks, Table.~\ref{Final Results} also contains the latency (in milliseconds). We can see that ResNet has the best accuracy of 99.26\%, but has a high latency of 117.28ms. InceptionV3 has an accuracy of 99.20\% which is close to ResNet, but with a lower latency of 94.1ms. We can see that InceptionV3, DenseNet, and ResNet have similar accuracies but the InceptionV3 model has the lowest latency. Thus, we chose InceptionV3 architecture.


\section{Conclusion}
In this paper, we designed an AI system for non-verbal communication that converts eye-blinks to words using a deep learning CNN architecture. The system predicts the state of the eyes of the patient and finds the blinking pattern. We compared several CNN architectures and hyperparameter selections in model training. For the evaluation, we tested our system using 16,979 facial images and found that our proposed prediction model was efficient and effective. Results demonstrate that overall prediction accuracy is 99.20\% and the average prediction time is 94ms. We trained different architectures with different hyperparameters to identify parameter combinations that lead to high accuracy and low latency. For the sake of conducting a clear comparative analysis, we compare the results of each architecture with batch sizes of 8, 16, and 32. SqueezeNet received the lowest accuracy with the fewest parameters. The DenseNet, ResNet, and InceptionV3 acquired accuracies in the range of 99.20\% and above. As InceptionV3 had the lowest latency, we chose this architecture. We introduced transfer learning, which improved the convergence when compared to random initialization, to otherwise similar accuracy and latency in the response. Future work includes the training on a more generalized training dataset, the application of hybrid technologies that fuse computer vision techniques such as the one presented together with other natural language processing methods, including recurrent neural networks, to introduce memory in the actions and system. The dataset, source code, demo, and results of this system are available at  \url{https://albara.ramli.net/research/bwcnn/}

%
%
%
%

\end{document}